# Remote Sensing Image Classification with Decoupled Knowledge Distillation

*Yaping He*[1], *Jianfeng Cai*[2], *Qicong Hu*[1*], *Peiqing Wang*[3]

[1]*College of Computer and Information Science College, Chongqing, China*
[2]*Chongqing Qingshan Industry Co., LTD, Chongqing, China*
[3]*Chengdu Jinhui Rongzhi Data Service Co., LTD, Chengdu, China*
**a1634688370@email.swu.edu.cn*

## Abstract

To address the challenges posed by the large number of parameters in existing remote sensing image classification models, which hinder deployment on resource-constrained devices, this paper proposes a lightweight classification method based on knowledge distillation. Specifically, G-GhostNet is adopted as the backbone network, leveraging feature reuse to reduce redundant parameters and significantly improve inference efficiency. In addition, a decoupled knowledge distillation strategy is employed, which separates target and non-target classes to effectively enhance classification accuracy. Experimental results on the RSOD and AID datasets demonstrate that, compared with the high-parameter VGG-16 model, the proposed method achieves nearly equivalent Top-1 accuracy while reducing the number of parameters by 6.24 times. This approach strikes an excellent balance between model size and classification performance, offering an efficient solution for deployment on resource-limited devices.

## 1 Introduction

With the rapid advancement of remote sensing technology, the volume of high-resolution remote sensing image data [1]-[10] has grown explosively, placing higher demands on the real-time processing capabilities of classification models [11]-[18]. However, traditional deep learning models often suffer from large parameter sizes and high computational complexity, making them difficult to deploy on resource-constrained edge devices or mobile platforms. Model lightweighting, by optimizing network structures, reducing parameter counts [19]-[33], and lowering computational overhead, can significantly improve inference efficiency while maintaining classification accuracy. This not only meets the requirements for real-time processing but also reduces hardware costs, making remote sensing image classification more accessible for practical applications such as agricultural monitoring, disaster assessment, and urban planning. In addition, lightweight models offer lower energy consumption, which is beneficial for building sustainable remote sensing application systems [34]-[43]. Therefore, advancing research on model lightweighting holds significant theoretical and practical value for promoting the real-world adoption and widespread use of remote sensing technologies [44]-[53].

In recent years, convolutional neural network (CNN)-based remote sensing image classification has been extensively studied, leading to the emergence of numerous research efforts. For example, Lei *et al*. [54] designed a capsule network with spectral-spatial features to enhance classification performance. Specifically, they employed 3D convolution to extend the routing mechanism, which reduced the number of parameters in the capsule layers, and further improved the network's performance by applying regularization through a deconvolution-based decoder. Shi *et al*. [55] proposed an adversarial network based on multi-feature collaboration, which efficiently solves the optimization problem by designing a customized loss function and employing an alternating optimization strategy. In addition, Burgert *et al*. [56] enhanced model robustness through regularization and joint collaborative regularized training. Zheng *et al*. [57] designed a hybrid architecture that integrates CNN and Transformer models. Specifically, convolutional layers are employed to extract multispectral spatial-temporal features, while the Transformer is used to capture global information, thereby enhancing the overall performance of the model. Li *et al*. [58] proposed a multi-task learning framework based on Transformers by adopting strategies such as shared encoding, achieving remarkable performance across classification, detection, and segmentation tasks. However, these approaches tend to overlook the balance between lightweight model design and classification accuracy, limiting their practical application in complex and resource-constrained scenarios.

Although lightweight networks can effectively reduce computational costs and improve model efficiency, a certain degree of accuracy loss is still inevitable. To further enhance the performance of student models, knowledge distillation [59]-[78] has been widely adopted to extract "dark knowledge" from teacher models, helping to compensate for this accuracy gap [79]-[88]. However, traditional knowledge distillation methods typically overlook the distinction between target and non-target classes, making it difficult to flexibly optimize the knowledge transfer process and limiting their adaptability to different learning tasks and datasets [89]-[95].

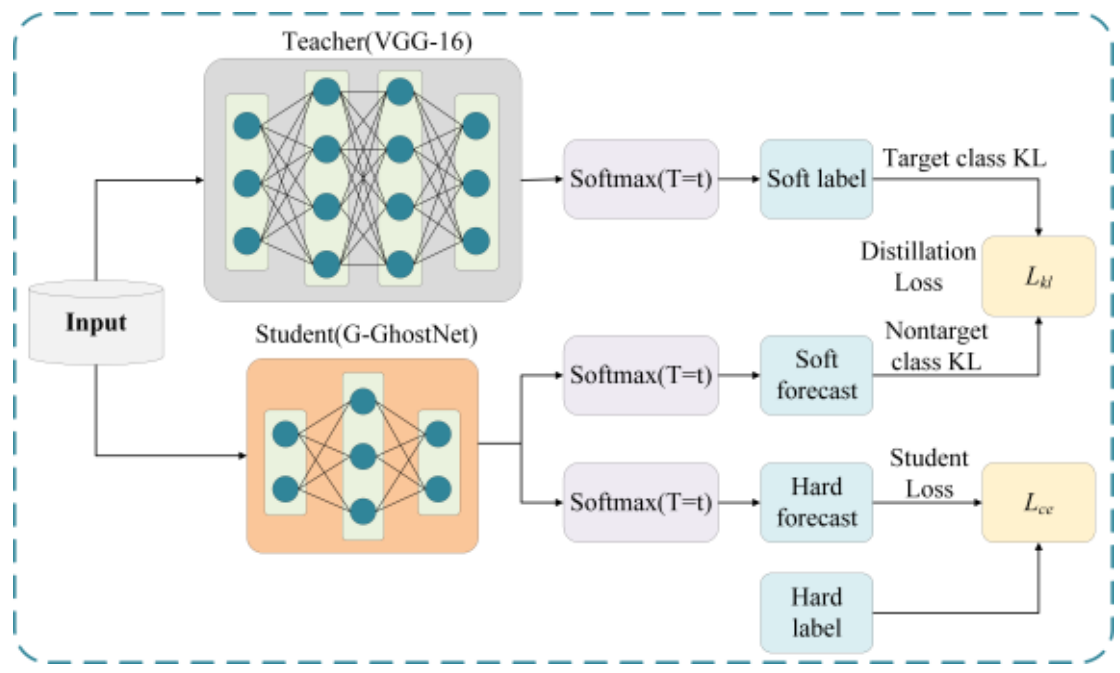

Fig. 1. Flowchart of remote sensing image classification

To address the above challenges, this paper proposes a remote sensing image classification model based on G-GhostNet and decoupled knowledge distillation. The main contributions of this work are as follows:

1) G-GhostNet is introduced as the backbone network to enhance the model's classification capability.
2) A decoupled knowledge distillation framework is adopted, which separates the loss between target and non-target classes, effectively compensating for the performance degradation caused by the lightweight network.
3) We evaluated the proposed framework on various remote sensing image classification datasets. Experimental results demonstrate that, compared to the high-parameter VGG-16 model, the proposed method achieves nearly equivalent Top-1 accuracy while reducing the number of parameters by a factor of 6.24.

The method proposed in this study demonstrates remarkable effectiveness and superiority in remote sensing image classification tasks. In summary, the model significantly reduces computational complexity and resource consumption while maintaining high classification accuracy, offering an efficient solution for real-time image processing in resource-constrained environments.

## 2. Methodology

In intelligent interpretation tasks of remote sensing imagery, classification models must address the dual challenges of computational intensity posed by high-resolution data and the real-time deployment requirements of edge devices. However, existing lightweight networks often face bottlenecks such as insufficient multi-scale feature extraction and inefficient spectral-spatial joint modeling in complex land cover scenarios. To overcome these limitations, this study proposes a knowledge distillation-based framework for remote sensing image classification. The model architecture is shown in Fig. 1.

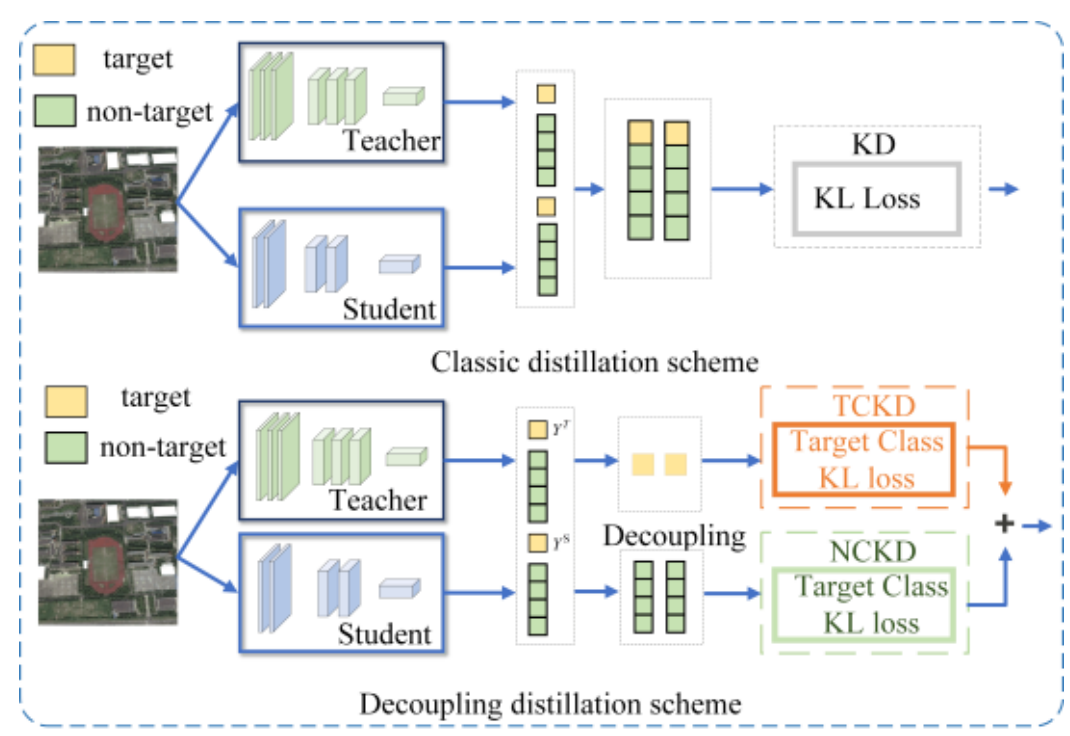

Fig. 2. Decoupled distillation architecture.

### 2.1 G-GhostNet

G-GhostNet is an improved version of GhostNet, designed specifically for lightweight visual tasks. Its core idea lies in reducing computational costs by leveraging feature redundancy. In remote sensing image classification scenarios, G-GhostNet effectively balances model accuracy and computational efficiency, making it particularly well-suited for handling high-resolution and large-scale remote sensing data. The complexity analysis is as follows: Give a stage consisting of $N$ blocks, the FLOPs of $i$th block are denoted as $f_i$ and the number of parameters as $p_i$. In the G-Ghost stage, the FLOPs of the original $n$ blocks are reduced to $f_1$ and $\{(1-\lambda)^2 f_i\}_{i=3}^{n}$. Similarly, the number of parameters is reduced to $p_1$, $(1-\lambda)\ p_2$, and $\{(1-\lambda)^2 f_i\}_{i=3}^{n}$. hence, the theoretical complexity is calculated as:

$$
\begin{cases}
r_f \approx \dfrac{\sum_{i=1}^{n} f_n}{f_1 + (1-\lambda) f_2 + \sum_{i=3}^{n} (1-\lambda)^2 f_n + f^c}, \\[2ex]
r_p \approx \dfrac{\sum_{i=1}^{n} p_n}{p_1 + (1-\lambda) p_2 + \sum_{i=3}^{n} (1-\lambda)^2 p_n + p^c},
\end{cases}
\tag{1}
$$

$r_p$ and $r_f$ respectively indicate the reduction ratios of the number of parameters and FLOPs. when $n \geqslant 3$, both the FLOPs and the number of parameters are significantly reduced. By adopting a standard CNN architecture, the G-GhostNet model structure can be easily derived.

Although G-GhostNet introduces substantial improvements over GhostNet, some degree of accuracy loss remains inevitable. Knowledge distillation, however, offers an effective solution to this issue. Therefore, this study integrates a decoupled distillation strategy to further enhance the performance of remote sensing image classification tasks.

*2.2 Knowledge distillation*

Knowledge distillation is a technique where a student model learns under the supervision of a teacher model. Traditional knowledge distillation methods usually assume that the logits of the teacher and student models have the same range and variance. However, this assumption is often difficult to satisfy in real-world applications, especially when there is a large capacity gap between the teacher and student models. To overcome this limitation, Decoupled Knowledge Distillation (DKD) has been proposed, as shown in Fig. 2. DKD improves the efficiency and accuracy of knowledge transfer by dividing the logits of both the teacher and student models into two parts, namely the target class and non-target classes, and performing distillation separately.

The Kullback-Leibler divergence is defined as follows:

$$
\begin{aligned}
L_{KD} &= \frac{T^2}{B} \sum_{i=1}^{B} \mathrm{KL}\left( Y_{i,:}^{(t,T)}, Y_{i,:}^{(s,T)} \right) \\
&= \frac{T^2}{B} \sum_{i=1}^{B} \sum_{j=1}^{N} Y_{ij}^{(t,T)} \log\left( \frac{Y_{ij}^{(t,T)}}{Y_{ij}^{(s,T)}} \right),
\end{aligned}
\tag{2}
$$

where $B$ denotes the batch size, $T$ represents the distillation temperature, and $\mathrm{Y}_{i,:}^{(t,T)}$ and $\mathrm{Y}_{i,:}^{(s,T)}$ refer to the outputs of the teacher model and the student model, respectively.

To enable the student model to fully learn both intra-class and inter-class relationships, Equation (2) can be decomposed into two parts.

$$
\begin{cases}
L_{\text{inter}} = \dfrac{T^2}{B} \sum_{i=1}^{B} \mathrm{KL}(Y_{i,:}^{(t,T)}, Y_{i,:}^{(s,T)}) \\[2ex]
L_{\text{intra}} = \dfrac{T^2}{N} \sum_{j=1}^{N} \mathrm{KL}(Y_{:,j}^{(t,T)}, Y_{:,j}^{(s,T)})
\end{cases}
\tag{3}
$$

Therefore, the original KD loss consists of $L_{\text{inter}}$ and $L_{\text{intra}}$:

$$
L_{\text{KD}} = L_{\text{inter}} + L_{\text{intra}} \tag{4}
$$

As a result, the final loss is defined as follows:

$$
L_{\text{total}} = L_{\text{task}} + L_{\text{KD}} \tag{5}
$$

where $L_{\text{task}}$ denotes the cross-entropy loss.

## 3 Experiment and Results

### *3.1 Experimental setup*

The experiments were conducted on a cloud server running the Linux operating system, with the network built using the Pytorch framework. The hardware configuration includes an RTX 3090 GPU (24GB), and the software environment consists of Pytorch 1.11.0 and CUDA 11.3. Stochastic Gradient Descent (SGD) was used for model optimization, with momentum and weight decay set to 0.937 and 0.00005, respectively. The training process consisted of 200 epochs, with a batch size of 32. The input image resolution was uniformly adjusted to 224×224.

### *3.2 Experimental dataset*

This study employed two widely used remote sensing image datasets: RSOD (Remote Sensing Object Detection) and AID (Aerial Image Dataset). To ensure the fairness of the experiments and the reliability of the results, the original datasets were randomly divided into training and testing sets at a ratio of 8:2.

### *3.3 Evaluation Metrics*

In image classification tasks, Top-1 Accuracy and Top-5 Accuracy are commonly used evaluation metrics to assess the classification performance of models on the test set. They are defined as follows:

$$\begin{cases} \text{Top-1 Accuracy} = \dfrac{N_{\text{correct}}}{N} \\ \text{Top-5 Accuracy} = \dfrac{N_{\text{top-5}}}{N} \end{cases} \tag{6}$$

Table 1. Ablation Study Results on the AID and RSOD Datasets

| Number | G-GhostNet | DKD | Params (M) | Top-1 (%) |
|---|---|---|---|---|
| 1 | × | × | 14.72 | 92.00/97.33 |
| 2 | ✓ | × | 2.36 | 89.10/94.76 |
| 3 | ✓ | ✓ | 2.36 | 91.65/96.85 |

$N_{\text{correct}}$ denotes the number of samples whose predicted class matches the ground truth, and $N$ represents the total number of samples. $N_{\text{top-5}}$ refers to the number of samples for which the true class appears among the top five predicted classes. In addition, the number of parameters and floating point operations (FLOPs) are also used as evaluation metrics to provide a more comprehensive assessment of the model's performance.

### *3.4 Ablation experiment*

Given the strong performance of VGG-16 on image classification tasks, it is adopted as the baseline model in this study. The results of the ablation experiments are presented in Table 1.

The ablation results demonstrate consistent trends across both datasets. Specifically, although the proposed model employs the lightweight G-GhostNet as the backbone, which significantly reduces the number of parameters, it still achieves competitive classification accuracy. Moreover, the application of decoupled knowledge distillation (DKD) leads to a noticeable improvement in performance, indicating the importance of separating the classification loss into target and non-target components. Additionally, the Logit normalization strategy further enhances the model's accuracy, suggesting a synergistic effect when combined with the decoupled knowledge distillation approach. These experimental results strongly confirm the effectiveness of the proposed method.

### *3.5 Performance Comparison of Different Classification Models*

To further demonstrate the advanced nature of the proposed method, several models including MobileNetV4 are compared in this paper. The comparison results are shown in Table 2.

On the RSOD dataset, the proposed method achieves a Top-1 accuracy of 91.29% and a Top-5 accuracy of 99.18% with only 2.36M parameters. Compared to DenseNet, despite DenseNet having fewer parameters (1.05M), the proposed method still outperforms it by 0.09 percentage points in Top-1 accuracy. When compared to the VGG-16 model, the proposed method reduces parameters significantly while maintaining comparable accuracy — with only a slight 0.35 percentage point drop in Top-1 accuracy, and a substantial reduction in FLOPs from 314.55M to just 4.35M, demonstrating its efficiency for resource-constrained devices.

Similarly, on the AID dataset, the proposed method achieves a Top-1 accuracy of 96.85%, which is comparable to EfficientNetB0 (97.20%) and surpasses MobileNetV4 (96.79%), while maintaining a much lower computational cost. For instance, the proposed method requires only 2.36M parameters and 4.35M FLOPs, compared to MobileNetV4's 2.97M parameters and 6.16M FLOPs. These results clearly demonstrate that the proposed method achieves competitive classification performance while significantly reducing model complexity, highlighting its suitability for lightweight deployment.

Table 2. Experimental Results of Comparative Study

| Model | Parameters (M) | Flops (M) | Top-1 (%) | Top-5 (%) |
|---|---|---|---|---|
| | | AID/RSOD | | |
| DenseNet | 1.05 | 292.81 | 91.56/96.79 | 99.17/- |
| EfficientNetB0 | 2.91 | 28.38 | 88.66/97.20 | 98.50/- |
| EfficientNetV2 | 38.17 | 80.56 | 89.29/92.30 | 99.05/- |
| GoogleNet | 6.16 | 1538.50 | 88.58/95.40 | 98.48/- |
| MobileNetV1 | 3.21 | 47.59 | 86.45/92.55 | 97.80/- |
| MobileNetV2 | 2.29 | 96.15 | 88.80/96.33 | 99.50/- |
| MobileNetV4 | 2.97 | 6.16 | 89.05/96.79 | 99.02/- |
| ShuffleNetv1 | 0.88 | 42.03 | 86.80/94.30 | 97.80/- |
| ShuffleNetv2 | 1.26 | 46.55 | 87.55/95.73 | 98.95/- |
| VGG-16 | 14.72 | 314.55 | 92.00/97.33 | 99.25/- |
| **Ours** | 2.36 | 4.35 | 91.65/96.85 | 99.18/- |

# 4 Conclusion

To address the challenge of deploying large-parameter models in resource-constrained environments, this paper proposes a knowledge distillation-based framework for remote sensing image classification. Specifically, a lightweight backbone, G-GhostNet, is introduced to accelerate inference while decoupled knowledge distillation is employed to compensate for potential accuracy loss. This strategy effectively mitigates the deployment bottleneck and demonstrates strong practical value. In future work, we plan to further explore constrained optimization from the perspectives of accuracy, parameter count, and computational cost, and introduce reinforcement learning-based approaches to build a quantization-aware framework that promotes efficient model deployment and application.

# 5 References

[1] S. Zhao, H. Chen, X. Zhang, P. Xiao, L. Bai and W. Ouyang, "RS-Mamba for Large Remote Sensing Image Dense Prediction," *IEEE Transactions on Geoscience and Remote Sensing*, vol. 62, pp. 1-14, 2024, Art no. 5633314.

[2] K. Chen, B. Chen, C. Liu, W. Li, Z. Zou and Z. Shi, "RSMamba: Remote Sensing Image Classification With State Space Model," *IEEE Geoscience and Remote Sensing Letters*, vol. 21, pp. 1-5, 2024, Art no. 8002605.

[3] N. Zeng, X. Li, P. Wu, H. Li and X. Luo, "A Novel Tensor Decomposition-Based Efficient Detector for Low-Altitude Aerial Objects With Knowledge Distillation Scheme," *IEEE/CAA J. Autom. Sinica*, vol. 11, no. 2, pp. 487-501, Feb. 2024.

[4] L. Wei, L. Jin and X. Luo, "Noise-Suppressing Neural Dynamics for Time-Dependent Constrained Nonlinear Optimization With Applications," *IEEE Trans. Syst. Man Cybern. Syst*., vol. 52, no. 10, pp. 6139-6150, Oct. 2022.

[5] L. Jin, Y. Qi, X. Luo, S. Li, and M. Shang, "Distributed Competition of Multi-Robot Coordination Under Variable and Switching Topologies," *IEEE Trans. Autom. Sci. Eng.*, vol. 19, no. 4, pp. 3575-3586, Oct. 2022.

[6] Y. He, Y. Su, X. Wang, J. Yu, and Y. Luo, "An improved method MSS-YOLOv5 for object detection with balancing speed-accuracy", *Front. Phys.*, vol. 10, pp. 1101923, 2023.

[7] Z. Liu, X. Luo and Z. Wang, "Convergence Analysis of Single Latent Factor-Dependent, Nonnegative, and Multiplicative Update-Based Nonnegative Latent Factor Models," *IEEE Trans. Neural Netw. Learn. Syst.*, vol. 32, no. 4, pp. 1737-1749, Apr. 2021.

[8] X. Luo, H. Wu, and Z. Li, "NeuLFT: A novel approach to nonlinear canonical polyadic decomposition on high-dimensional incomplete tensors," *IEEE Trans. Knowl. Data Eng*., vol. 35, no. 6, pp. 6148-6166, Jun. 2023.

[9] X. Luo, D. Wang, M. Zhou, and H. Yuan, "Latent Factor-Based Recommenders Relying on Extended Stochastic Gradient Descent Algorithms," *IEEE Trans. Syst. Man Cybern. Syst*., vol. 51, no. 2, pp. 916-926, Feb. 2021.

[10] S. Deng, Q. Hu, D. Wu, and Y. He, "BCTC-KSM: A blockchain-assisted threshold cryptography for key security management in power IoT data sharing," *Comput. Electr. Eng*., vol. 108, pp: 108666, 2023.

[11] Y. Xiao, Q. Yuan, K. Jiang, J. He, X. Jin, and L. Zhang, "EDiffSR: An Efficient Diffusion Probabilistic Model for Remote Sensing Image Super-Resolution," *IEEE Transactions on Geoscience and Remote Sensing*, vol. 62, pp. 1-14, 2024, Art no. 5601514.

[12] X. Luo, W. Qin, A. Dong, K. Sedraoui and M. Zhou, "Efficient and High-quality Recommendations via Momentum-incorporated Parallel Stochastic Gradient Descent-Based Learning," *IEEE/CAA J. Autom. Sin.*, vol. 8, no. 2, pp. 402-411, Feb. 2021.

[13] W. Li, X. Luo, H. Yuan, and M. Zhou, "A Momentum-Accelerated Hessian-Vector-Based Latent Factor Analysis Model," *IEEE Trans. Serv. Comput*., vol. 16, no. 2, pp. 830-844, 1 Mar.-Apr. 2023.

[14] M.Chen, C. He, and X. Luo, “MNL: A Highly-Efficient model for Large-scale Dynamic Weighted Directed Network Representation,” *IEEE Trans. on Big Data*, vol. 9, no. 3, pp. 889-903, Oct. 2023.

[15] X. Luo, H. Wu, Z. Wang, J. Wang, and D. Meng, “A Novel Approach to Large-Scale Dynamically Weighted Directed Network Representation,” *IEEE Trans. Pattern Anal. Mach. Intell*., vol. 44, no. 12, pp. 9756-9773, 1 Dec. 2022.

[16] J. Chen, X. Luo and M. Zhou, “Hierarchical Particle Swarm Optimization-incorporated Latent Factor Analysis for Large-Scale Incomplete Matrices,” *IEEE Trans. Big Data*, vol. 8, no. 6, pp. 1524-1536, 1 Dec. 2022.

[17] L. Hu, S. Yang, X. Luo, H. Yuan, K. Sedraoui, and M. Zhou, “A Distributed Framework for Large-scale Protein-protein Interaction Data Analysis and Prediction Using MapReduce,” *IEEE/CAA J. Autom. Sin.*, vol. 9, no. 1, pp. 160-172, Jan. 2022.

[18] Z. Wang, Y. Liu, X. Luo, J. Wang, C. Gao, D. Peng, and W.Chen, “Large-Scale Affine Matrix Rank Minimization With a Novel Nonconvex Regularizer,” *IEEE Trans. Neural Networks Learn. Syst*., vol. 33, no. 9, pp. 4661-4675, Sept. 2022.

[19] X. Liao, K. Hoang and X. Luo, “Local Search-Based Anytime Algorithms for Continuous Distributed Constraint Optimization Problems,” *IEEE/CAA Journal of Automatica Sinica*, vol. 12, no. 1, pp. 288-290, Jan. 2025.

[20] Y. He, X. Luo, “Tensor Low-Rank Orthogonal Compression for Convolutional Neural Networks,” IEEE/CAA Journal of Automatica Sinica, 2025, doi: 10.1109/JAS.2025.125213.

[21] M. Chen, R. Wang, Y. Qiao, and X. Luo, “A Generalized Nesterov's Accelerated Gradient-Incorporated Non-Negative Latent-Factorization-of-Tensors Model for Efficient Representation to Dynamic QoS Data,” *IEEE Trans. Emerg. Top. Comput. Intell*., vol. 8, no. 3, pp. 2386-2400, Jun. 2024.

[22] Q. Wang and H. Wu, “Dynamically Weighted Directed Network Link Prediction Using Tensor Ring Decomposition,” *27th Int. Conf. Comput. Supported Cooperative Work Des*, 2024, Tianjin, China, pp. 2864-2869.

[23] X. Luo, M. Chen, H. Wu, Z. Liu, H. Yuan, and M. Zhou, “Adjusting Learning Depth in Nonnegative Latent Factorization of Tensors for Accurately Modeling Temporal Patterns in Dynamic QoS Data,” *IEEE Trans. Autom. Sci. Eng*., vol. 18, no. 4, pp. 2142-2155, Oct. 2021.

[24] H. Wu, X. Luo, M. Zhou, M. J. Rawa, K. Sedraoui, and A. Albeshri, “A PID-incorporated Latent Factorization of Tensors Approach to Dynamically Weighted Directed Network Analysis,” *IEEE/CAA J. Autom. Sin.*, vol. 9, no. 3, pp. 533-546, Mar. 2022.

[25] W. Qin, H. Wang, F. Zhang, J. Wang, X. Luo and T. Huang, “Low-Rank High-Order Tensor Completion With Applications in Visual Data,” *IEEE Trans. Image Process*., vol. 31, pp. 2433-2448, Mar. 2022.

[26] H. Wu, X. Wu, and X. Luo, “Dynamic Network Representation Based on Latent Factorization of Tensors,” Springer, Mar. 2023.

[27] W. Li, X. Luo, H. Yuan, and M. Zhou, “A Momentum-Accelerated Hessian-Vector-Based Latent Factor Analysis Model,” *IEEE Trans. Serv. Comput*., vol. 16, no. 2, pp. 830-844, 1 Mar.-Apr. 2023.

[28] D. Wu, X. Luo, Y. He, and M. Zhou, “A Prediction-Sampling-Based Multilayer-Structured Latent Factor Model for Accurate Representation to High-Dimensional and Sparse Data,” *IEEE Trans. Neural Networks Learn. Syst*., vol. 35, no. 3, pp. 3845-3858, Mar. 2024.

[29] X. Luo, Y. Zhou, Z. Liu, and M. Zhou, “Fast and Accurate Non-Negative Latent Factor Analysis of High-Dimensional and Sparse Matrices in Recommender Systems,” *IEEE Trans. Knowl. Data Eng.*, vol. 35, no. 4, pp. 3897-3911, Apr. 2023.

[30] J. Wang, H. Wu and C. He, “A Double-Norm Aggregated Latent Factorization of Tensors Model for Temporal-Aware QoS Prediction,” Conf. Proc. *IEEE Int. Conf. Syst. Man Cybern*., 2023 , pp. 3913-3918.

[31] D. Wu, Y. He and X. Luo, “A Graph-Incorporated Latent Factor Analysis Model for High-Dimensional and Sparse Data,” *IEEE Trans. Emerg. Top. Comput.*, vol. 11, no. 4, pp. 907-917, Oct.-Dec. 2023.

[32] Y. Yuan, X. Luo, M. Shang, and Z. Wang, “A Kalman-Filter-Incorporated Latent Factor Analysis Model for Temporally Dynamic Sparse Data,” *IEEE Trans. Syst. Man Cybern. Syst*., vol. 53, no. 9, pp. 5788-5801, Sept. 2023.

[33] M. Chen and H. Wu, “A Breif Review on Data-driven Battery Health Estimation Methods for Energy Storage Systems,” *Conf. Proc. IEEE Int. Conf. Syst. Man Cybern*., 2022, pp. 1-6.

[34] Y. Yuan, Q. He, X. Luo and M. Shang, “A Multilayered-and-Randomized Latent Factor Model for High-Dimensional and Sparse Matrices,” *IEEE Trans. Big Data*, vol. 8, no. 3, pp. 784-794, 1 June 2022.

[35] D. Wu, Y. He, X. Luo and M. Zhou, “A Latent Factor Analysis-Based Approach to Online Sparse Streaming Feature Selection,” *IEEE Trans. Syst. Man Cybern. Syst., vol*. 52, no. 11, pp. 6744-6758, Nov. 2022.

[36] Y. He and L. Xiao, “Structured Pruning for Deep Convolutional Neural Networks: A Survey,” *IEEE Trans. Pattern Anal. Mach. Intell*., vol. 46, no. 5, pp. 2900-2919, May 2024.

[37] H. Wu, Y. Qiao and X. Luo, “A Fine-Grained Regularization Scheme for Non-negative Latent Factorization of High-Dimensional and Incomplete Tensors,” *IEEE Trans. Serv. Comput.*, vol. 17, no. 6, pp. 3006-3021, Nov.-Dec. 2024.

[38] W. Qin, X. Luo and M. Zhou, “Adaptively-Accelerated Parallel Stochastic Gradient Descent for High-Dimensional and Incomplete Data Representation Learning,” *IEEE Trans. Big Data*, vol. 10, no. 1, pp. 92-107, Feb. 2024.

[39] W. Qin and X. Luo, “Asynchronous Parallel Fuzzy Stochastic Gradient Descent for High-Dimensional Incomplete Data Representation,” *IEEE Trans. Fuzzy Syst*., vol. 32, no. 2, pp. 445-459, Feb. 2024.

[40] H. Wu, X. Luo and M. Zhou, “Advancing Non-Negative Latent Factorization of Tensors With Diversified Regularization Schemes,” *IEEE Trans. Serv*., vol. 15, no. 3, pp. 1334-1344, 1 May-June 2022.

[41] X. Luo, M. Chen, H. Wu, Z. Liu, H. Yuan and M. Zhou, “Adjusting Learning Depth in Nonnegative Latent Factorization of Tensors for Accurately Modeling Temporal Patterns in Dynamic QoS Data,” *IEEE Trans. Autom. Sci. Eng*., vol. 18, no. 4, pp. 2142-2155, Oct. 2021.

[42] M. Chen, H. Wu, C. He and S. Chen, “Momentum-incorporated Latent Factorization of Tensors for Extracting Temporal Patterns from QoS Data,” 2019 *Conf. Proc. IEEE Int. Conf. Syst. Man Cybern*., 2019, pp. 1757-1762.

[43] M. Yin, Y. Sui, W. Yang, X. Zang, Y. Gong, and B. Yuan, “HODEC: Towards Efficient High-order Decomposed Convolutional Neural Networks,” in *Proc. 35th IEEE/CVF Conf. Comput. Vis. Pattern Recognit.*, New Orleans, USA, 2022, pp. 12299-12308.

[44] L. Jin, X. Zheng and X. Luo, “Neural Dynamics for Distributed Collaborative Control of Manipulators With Time Delays,” IEEE/CAA J. Autom. Sin., vol. 9, no. 5, pp. 854-863, May 2022.

[45] W. Yang, S. Li, Z. Li and X. Luo, “Highly Accurate Manipulator Calibration via Extended Kalman Filter-Incorporated Residual Neural Network,” IEEE Trans. Ind. Informat., vol. 19, no. 11, pp. 10831-10841, Nov. 2023.

[46] L. Chen and X. Luo, “Tensor Distribution Regression Based on the 3D Conventional Neural Networks,” IEEE/CAA J. Autom. Sin., vol. 10, no. 7, pp. 1628-1630, Jul. 2023.

[47] L. Wei, L. Jin and X. Luo, “Noise-Suppressing Neural Dynamics for Time-Dependent Constrained Nonlinear Optimization With Applications,” IEEE Trans. Syst. Man Cybern. Syst., vol. 52, no. 10, pp. 6139-6150, Oct. 2022.

[48] Y, Zheng, T. Huang, X. Zhao, Q. Zhao, and T. Jiang, "Fully-Connected Tensor Network Decomposition and Its Application to Higher-Order Tensor Completion," 35th Proc. AAAI Conf. Artif. Intell., 2021, Online, vol.35, no. 12, pp: 4683-4690.

[49] V. Aggarwal, W. Wang, B. Eriksson, Y. Sun, and W. Wang, "Wide Compression: Tensor Ring Nets," in Proc. 31st IEEE/CVF Conf. Comput. Vis. Pattern Recognit., Hawaii, USA, 2018, pp. 9329-9338.

[50] H. Wu and X. Luo, "Instance-Frequency-Weighted Regularized, Nonnegative and Adaptive Latent Factorization of Tensors for Dynamic QoS Analysis," IEEE Int. Conf. Web Serv., 2021, Chicago, USA, 2021, pp. 560-568.

[51] H. Wu, X. Luo and M. Zhou, "Neural Latent Factorization of Tensors for Dynamically Weighted Directed Networks Analysis," IEEE Int. Conf. Syst. Man Cybern., 2021, Melbourne, Australia, pp. 3061-3066.

[52] D. Wu, P. Zhang, Y. He, and X. Luo, "A Double-Space and Double-Norm Ensembled Latent Factor Model for Highly Accurate Web Service QoS Prediction," IEEE Trans. Serv. Comput., vol. 16, no. 2, pp. 802-814, 1 Mar.-Apr. 2023.

[53] Z. Xie, L. Jin and X. Luo, "Kinematics-Based Motion-Force Control for Redundant Manipulators With Quaternion Control," IEEE Trans. Autom. Sci. Eng., vol. 20, no. 3, pp. 1815-1828, Jul. 2023.

[54] R. Lei, C. Zhang, W. Liu, L. Zhang, X. Zhang, and Y. Yang, "Hyperspectral Remote Sensing Image Classification Using Deep Convolutional Capsule Network," *IEEE Journal of Selected Topics in Applied Earth Observations and Remote Sensing*, vol. 14, pp. 8297-8315, 2021.

[55] C. Shi, Y. Dang, L. Fang, M. Zhao, Z. Lv, and Q. Miao, "Multifeature Collaborative Adversarial Attack in Multimodal Remote Sensing Image Classification," *IEEE Transactions on Geoscience and Remote Sensing*, vol. 60, pp. 1-15, 2022, Art no. 5631815.

[56] T. Burgert, M. Ravanbakhsh and B. Demir, "On the Effects of Different Types of Label Noise in Multi-Label Remote Sensing Image Classification," *IEEE Transactions on Geoscience and Remote Sensing*, vol. 60, pp. 1-13, 2022, Art no. 5413713.

[57] Y. Zheng, S. Liu, H. Chen and L. Bruzzone, "Hybrid FusionNet: A Hybrid Feature Fusion Framework for Multisource High-Resolution Remote Sensing Image Classification," *IEEE Transactions on Geoscience and Remote Sensing*, vol. 62, pp. 1-14, 2024, Art no. 5401714.

[58] Q. Li, Y. Chen, X. He and L. Huang, "Co-Training Transformer for Remote Sensing Image Classification, Segmentation, and Detection," in IEEE Transactions on Geoscience and Remote Sensing, vol. 62, pp. 1-18, 2024, Art no. 5606218.

[59] W. Qin, H. Wu, Q. Lai and C. Wang, "A Parallelized, Momentum-incorporated Stochastic Gradient Descent Scheme for Latent Factor Analysis on High-dimensional and Sparse Matrices from Recommender Systems," *Conf. Proc. IEEE Int. Conf. Syst. Man Cybern*., 2019, pp. 1744-1749.

[60] Y. Yuan, J. Li, and X. Luo, "A Fuzzy PID-Incorporated Stochastic Gradient Descent Algorithm for Fast and Accurate Latent Factor Analysis," *IEEE Trans. Fuzzy Syst.*, vol. 32, no. 7, pp. 4049-4061, 2024.

[61] J. Wang, W. Li, and X. Luo, "A Distributed Adaptive Second-Order Latent Factor Analysis Model," *IEEE/CAA J. Autom. Sin*., vol. 11, no. 11, pp. 2343-2345, 2024.

[62] Y. Song, M. Li, X. Luo, G. Yang, and C. Wang, "Improved Symmetric and Nonnegative Matrix Factorization Models for Undirected, Sparse and Large-Scaled Networks: A Triple Factorization-Based Approach," *IEEE Trans. Ind. Inform*., vol. 16, no. 5, pp. 3006-3017, 2020.

[63] X. Luo, Z. Liu, S. Li, M. Shang, and Z. Wang, "A Fast Non-Negative Latent Factor Model Based on Generalized Momentum Method," *IEEE Trans. Syst. Man Cybern. Syst.*, vol. 51, no. 1, pp. 610-620, 2021.

[64] X. Luo, M. Zhou, S. Li, and M. Shang, "An Inherently Nonnegative Latent Factor Model for High-Dimensional and Sparse Matrices from Industrial Applications," *IEEE Trans. Ind. Inform*. vol. 14, no. 5, pp. 2011-2022, 2018.

[65] Y. Song, M. Li, Z. Zhu, G. Yang, and X. Luo, "Nonnegative Latent Factor Analysis-Incorporated and Feature-Weighted Fuzzy Double c-Means Clustering for Incomplete Data," *IEEE Trans. Fuzzy Syst.*, vol. 30, no. 10, pp. 4165-4176, 2022.

[66] Z. Liu, G. Yuan, and X. Luo, "Symmetry and Nonnegativity-Constrained Matrix Factorization for Community Detection," *IEEE/CAA J. Autom. Sin*., vol. 9, no. 9, pp. 1691-1693, 2022.

[67] Y. Yang, X. Sun, W. Diao, H. Li, Y. Wu, and X. Li, "Adaptive Knowledge Distillation for Lightweight Remote Sensing Object Detectors Optimizing," *IEEE Transactions on Geoscience and Remote Sensing*, vol. 60, pp. 1-15, 2022, Art no. 5623715.

[68] C. K. Joshi, F. Liu, X. Xun, J. Lin and C. S. Foo, "On Representation Knowledge Distillation for Graph Neural Networks," *IEEE Transactions on Neural Networks and Learning Systems*, vol. 35, no. 4, pp. 4656-4667, April 2024.

[69] W. Li, R. Wang, X. Luo and M. Zhou, "A Second-Order Symmetric Non-Negative Latent Factor Model for Undirected Weighted Network Representation," *IEEE Trans. Netw. Sci. Eng*., vol. 10, no. 2, pp. 606-618, 1 Mar.-Apri. 2023.

[70] Y. Yuan, R. Wang, G. Yuan and L. Xin, "An Adaptive Divergence-Based Non-Negative Latent Factor Model," *IEEE Trans. Syst. Man Cybern. Syst*., vol. 53, no. 10, pp. 6475-6487, Oct. 2023.

[71] C. Li, C.shi, "Constrained Optimization Based Low-Rank Approximation of Deep Neural Networks," in *31st Proc. Eur. Conf. Comput. Vis*., 2018, Salt Lake City, USA, pp:732-747.

[72] X. Luo, H. Wu, H. Yuan and M. Zhou, "Temporal Pattern-Aware QoS Prediction via Biased Non-Negative Latent Factorization of Tensors," *IEEE Trans. Cybern*., vol. 50, no. 5, pp. 1798-1809, May 2020.

[73] W. Deng, X. Zhang, Y. Zhou, Y. Liu, X. Zhou, H. Chen, and H. Zhao, "An enhanced fast non-dominated solution sorting genetic algorithm for multi-objective problems," *Inf. Sci*., vol. 585, pp. 1-13, Mar. 2021.

[74] M. Beniwal, A. Singh, N. Kumar, "Forecasting long-term stock prices of global indices: A forward-validating Genetic Algorithm optimization approach for Support Vector Regression," *Appl. Soft Comput.*, vol. 145, pp. 1-15, Mar. 2023.

[75] T. Chen, S. Li, Y. Qiao, and X. Luo, "A Robust and Efficient Ensemble of Diversified Evolutionary Computing Algorithms for Accurate Robot Calibration," *IEEE Trans. Instrum. Meas*., vol. 73, pp. 1-14, Feb. 2024.

[76] H. Wu, X. Luo and M. Zhou, "Discovering Hidden Pattern in Large-scale Dynamically Weighted Directed Network via Latent Factorization of Tensors," *IEEE 17th Int. Conf. Autom. Sci. Eng*., 2021, Lyon, France, pp. 1533-1538.

[77] H. Wu, Y. Xia and X. Luo, "Proportional-Integral-Derivative-Incorporated Latent Factorization of Tensors for Large-Scale Dynamic Network Analysis," *China Autom. Congr*., 2021, Beijing, China, pp. 2980-2984.

[78] L. Acerbi, "Variational Bayesian Monte Carlo with Noisy Likelihoods," *33rd Adv. Neural Inf. Process. Syst*., 2020, Online, pp: 8211-8222.

[79] F. Bi, X. Luo, B. Shen, H. Dong, and Z. Wang, "Proximal Alternating-Direction-Method-of-Multipliers-Incorporated Nonnegative Latent Factor Analysis," *IEEE/CAA J. Autom. Sin.*, vol. 10, no. 6, pp. 1388-1406, Jun. 2023.

[80] Y. Zhong, K. Liu, S. Gao, and X. Luo, "Alternating-Direction-Method of Multipliers-Based Adaptive Nonnegative Latent Factor Analysis," *IEEE Trans. Emerg. Top. Comput. Intell*., doi: 10.1109/TETCI.2024.3420735.

[81] X. Luo, Y. Zhong, Z. Wang, and M. Li, "An Alternating-Direction-Method of Multipliers-Incorporated Approach to Symmetric Non-Negative Latent Factor Analysis," *IEEE Trans. Neural Netw. Learn. Syst.*, vol. 34, no. 8, pp. 4826-4840, Aug. 2023.

[82] W. Deng, J. Xu, X. -Z. Gao and H. Zhao, "An Enhanced MSIQDE Algorithm With Novel Multiple Strategies for Global Optimization Problems," *IEEE Trans. Syst. Man Cybern. Syst*., vol. 52, no. 3, pp. 1578-1587, March 2022.

[83] L. Hu, Y. Yang, Z. Tang, Y. He, and X. Luo, "FCAN-MOPSO: An Improved Fuzzy-Based Graph Clustering Algorithm for Complex Networks With Multiobjective Particle Swarm Optimization," *IEEE Trans. Fuzzy Syst*., vol. 31, no. 10, pp. 3470-3484, Oct. 2023.

[84] X. Luo, J. Chen, Y. Yuan, and Z. Wang, "Pseudo Gradient-Adjusted Particle Swarm Optimization for Accurate Adaptive Latent Factor Analysis," *IEEE Trans. Syst. Man Cybern. Syst*., vol. 54, no. 4, pp. 2213-2226, Apr. 2024.

[85] X. Sun, Z. Shi, G. Lei, Y. Guo, and J. Zhu, "Multi-Objective Design Optimization of an IPMSM Based on Multilevel Strategy," *IEEE Trans. Ind. Electron*., vol. 68, no. 1, pp. 139-148, Jan. 2021.

[86] C. He, R. Cheng, and D. Yazdani, "Adaptive Offspring Generation for Evolutionary Large-Scale Multiobjective Optimization," *IEEE Trans. Syst. Man Cybern. Syst*., vol. 52, no. 2, pp. 786-798, Feb. 2022.

[87] Z. Xie, L. Jin, X. Luo, M. Zhou, and Y. Zheng, "A Biobjective Scheme for Kinematic Control of Mobile Robotic Arms With Manipulability Optimization," *IEEE/ASME Trans. Mechatron*., vol. 29, no. 2, pp. 1534-1545, Apr. 2024.

[88] X. Luo, Y. Yuan, S. Chen, N. Zeng, and Z. Wang, "Position-Transitional Particle Swarm Optimization-Incorporated Latent Factor Analysis," *IEEE Trans. Knowl. Data Eng*., vol. 34, no. 8, pp. 3958-3970, Aug. 2022.

[89] X. Xu, M. Lin, X. Luo, and Z. Xu, "HRST-LR: A Hessian Regularization Spatio-Temporal Low Rank Algorithm for Traffic Data Imputation," *IEEE Trans. Intell. Transp. Syst.*, vol. 24, no. 10, pp. 11001-11017, Oct. 2023.

[90] D. Wu, Q. He, X. Luo, M. Shang, Y. He, and G. Wang, "A Posterior-Neighborhood-Regularized Latent Factor Model for Highly Accurate Web Service QoS Prediction," *IEEE Trans. Serv. Comput*., vol. 15, no. 2, pp. 793-805, 2019.

[91] Z. Xie, L. Jin, X. Luo, S. Li, and X. Xiao, "A Data-Driven Cyclic-Motion Generation Scheme for Kinematic Control of Redundant Manipulators," *IEEE Trans. Control Syst. Technol*., vol. 29, no. 1, pp. 53-63, 2021.

[92] L. Hu, S. Yang, X. Luo, and M. Zhou, "An Algorithm of Inductively Identifying Clusters From Attributed Graphs," *IEEE Trans. Big Data*, vol. 8, no. 2, pp. 523-534, 2022.

[93] D. Wu, X. Luo, M. Shang, Y. He, G. Wang, and M. Zhou, "A Deep Latent Factor Model for High-Dimensional and Sparse Matrices in Recommender Systems," *IEEE Trans. Syst. Man Cybern. Syst*., vol. 51, no. 7, pp. 4285-4296, 2021.

[94] J. Chen, R. Wang, D. Wu, and X. Luo, "A Differential Evolution-Enhanced Position-Transitional Approach to Latent Factor Analysis," *IEEE Trans. Emerg. Top. Comput. Intell*., vol. 7, no. 2, pp. 389-401, 2023.

[95] F. Bi, T. He, and X. Luo, "A Fast Nonnegative Autoencoder-Based Approach to Latent Feature Analysis on High-Dimensional and Incomplete Data," *IEEE Trans. Serv. Comput.*, vol. 17, no. 3, pp. 733-746, 2024.